\documentclass[a4paper,fleqn]{cas-dc}

\usepackage[numbers]{natbib}

\def\tsc#1{\csdef{#1}{\textsc{\lowercase{#1}}\xspace}}
\tsc{WGM}
\tsc{QE}
\tsc{EP}
\tsc{PMS}
\tsc{BEC}
\tsc{DE}

\begin{document}
\let\WriteBookmarks\relax
\def\floatpagefraction{1}

\shorttitle{Variational Garrote for Sparse Inverse Problems}

\shortauthors{K Lee et~al.}

\title [mode = title]{Variational Garrote for Sparse Inverse Problems}



%
\author[1]{Kanghun Lee}


\affiliation[1]{
    organization={Department of Science Education, Seoul National University},
    city={Seoul},
    postcode={08826}, 
    country={Korea}}

\author[1]{Hyungjoon Soh}
\cormark[1]
\ead{hjsoh88@snu.ac.kr}

\author[1,2,3]{Junghyo Jo}
\cormark[1]
\ead{jojunghyo@snu.ac.kr}


\affiliation[2]{
    organization={Center for Theoretical Physics and Artificial Intelligence Institute, Seoul National University},
    city={Seoul},
    postcode={08826}, 
    country={Korea}}

\affiliation[3]{
    organization={School of Computational Sciences, Korea Institute for Advanced Study},
    city={Seoul},
    postcode={02455}, 
    country={Korea}}

\cortext[cor1]{Corresponding author}

\begin{abstract}
Sparse regularization plays a central role in solving inverse problems arising from incomplete or corrupted measurements. Different regularizers correspond to different prior assumptions about the structure of the unknown signal, and reconstruction performance depends on how well these priors match the intrinsic sparsity of the data. This work investigates the effect of sparsity priors in inverse problems by comparing conventional $\ell_1$ regularization with the Variational Garrote (VG), a probabilistic method that approximates $\ell_0$ sparsity through variational binary gating variables. A unified experimental framework is constructed across multiple reconstruction tasks including signal resampling, signal denoising, and sparse-view computed tomography. To enable consistent comparison across models with different parameterizations, regularization strength is swept across wide ranges and reconstruction behavior is analyzed through train–generalization error curves. Experiments reveal characteristic bias–variance tradeoff patterns across tasks and demonstrate that VG frequently achieves lower minimum generalization error and improved stability in strongly underdetermined regimes where accurate support recovery is critical. These results suggest that sparsity priors closer to spike-and-slab structure can provide advantages when the underlying coefficient distribution is strongly sparse. The study highlights the importance of prior–data alignment in sparse inverse problems and provides empirical insights into the behavior of variational $\ell_0$-type methods across different information bottlenecks.
\end{abstract}



\maketitle

\section{Introduction}

Inverse problems arise throughout signal processing and computational imaging, where the objective is to reconstruct an unknown signal or image from incomplete, indirect, or corrupted measurements~\cite{Benning2018}. Representative examples include missing-sample interpolation, denoising, deblurring, and tomographic reconstruction based on the Radon transform~\cite{Radon1917}. In these settings, measurements are produced by a forward process that reduces or distorts the information available about the original object.

Reconstruction from such measurements is frequently ill-posed: subsampling discards observations, projection operators aggregate information along measurement paths, and noise corrupts the acquired data. Successful reconstruction therefore requires additional structural assumptions beyond data fidelity.

A widely used assumption is sparsity. Although many signals are not sparse in their raw representation, they often admit sparse or compressible representations in transform domains such as the Fourier transform~\cite{Gilbert2014}, wavelet transform~\cite{Shensa1992}, or discrete cosine transform (DCT)~\cite{Gupta1990,Stankovic2018}. This principle underlies compressed sensing, which shows that sparse-structured signals can be recovered from far fewer measurements than predicted by classical sampling limits under suitable conditions~\cite{Shannon1949,Donoho2006,Candes2006,Candes2008}, and has motivated a broad range of reconstruction methods and applications~\cite{Lustig2007,Knoll2020,Duarte2008,Rousset2017,Haneche2021,Xu2022SparseSAR}.

In practice, sparsity is enforced through regularization~\cite{Benning2018}. From a Bayesian viewpoint, a regularizer corresponds to an implicit prior distribution on unknown coefficients: $\ell_1$ penalties correspond to Laplace priors, while ideal sparse recovery corresponds to $\ell_0$ formulations associated with spike-and-slab type priors that explicitly separate active and inactive coefficients~\cite{ref:NNG, ref:SpikeAndSlab, George1997Approaches}. The dominant convex approach is $\ell_1$ regularization, most notably the Least Absolute Shrinkage and Selection Operator (LASSO)~\cite{Tibshirani1996, Park2008}. Despite its tractability and strong theory, $\ell_1$ regularization imposes continuous shrinkage that can bias large coefficients and yield unstable selection under correlated predictors~\cite{Zou2005,Zhang2008}, often producing approximate rather than decisive sparsity.

The Variational Garrote (VG)~\cite{Kappen2014} provides a tractable approximation to $\ell_0$ sparsity via latent binary gating variables and a variational relaxation. By decoupling support selection from coefficient magnitude estimation, VG behaves similarly to spike-and-slab models while retaining a single differentiable objective amenable to gradient-based optimization. Recent analysis further suggests that VG can exhibit sharp transition behavior and robust support recovery in sparse regimes~\cite{Soh2025}. However, compared with $\ell_1$-based reconstruction, VG has been less systematically evaluated across practical inverse problems with diverse forward operators and information bottlenecks.

This work focuses on how sparsity priors influence reconstruction under controlled information bottlenecks by comparing LASSO and VG across three representative tasks: sound signal resampling, sound signal denoising, and sparse-view CT reconstruction. To enable model-agnostic comparison despite incomparable hyperparameters, we analyze train--generalization error curves obtained by sweeping regularization strength over a wide range, and compare methods at their best achievable generalization performance under each bottleneck.

Across both signal and image reconstruction, VG frequently achieves lower minimum generalization error and improved stability in strongly underdetermined regimes, consistent with the benefit of priors closer to spike-and-slab sparsity when the underlying coefficient distribution is strongly sparse~\cite{Soh2025}. The remainder of the paper is organized as follows. Section~2 introduces the unified sparse reconstruction formulation and VG. Section~3 describes datasets and the experimental protocol. Section~4 presents empirical results, and Section~5 concludes with implications and future directions.

\section{Theory}

\begin{table*}[b]
\centering
\small
\setlength{\tabcolsep}{6pt}
\renewcommand{\arraystretch}{1.15}
\begin{tabular}{llll}
\hline
\textbf{Task} & \textbf{Forward Operator $\mathbf{A}$} & \textbf{Information Bottleneck} & \textbf{Sparsity Control} \\
\hline
Signal Resampling & Subsampling / masking operator & Sampling ratio $R=M/N$ & transform sparsity (DCT) \\
Signal Denoising  & Identity ($\mathbf{A}=\mathbf{I}$) & Noise amplitude $\alpha$ & transform sparsity (DCT) \\
Sparse-view CT Reconstruction & Discrete Radon transform & Number of angles $K$ & image-domain sparsity \\
\hline
\end{tabular}
\caption{Inverse problems considered in this work expressed within a unified sparse reconstruction framework. All tasks share the same sparse regression formulation but differ in the forward operator and the form of information bottleneck.}
\label{tab:tasks}
\end{table*}

\subsection{Regularized Inverse Problems}

Many signal reconstruction tasks can be formulated as linear inverse problems. Let $\mathbf{x}\in\mathbb{R}^{N}$ denote an unknown signal (or vectorized image) and $\mathbf{y}\in\mathbb{R}^{M}$ denote observations generated by a forward operator $\mathbf{A}$,

\begin{equation}
    \mathbf{y} = \mathbf{A}\mathbf{x} + \boldsymbol{\epsilon},
\end{equation}
where $\boldsymbol{\epsilon}$ represents noise or modeling error.

When the system is underdetermined ($M\ll N$) or when $\mathbf{A}$ is ill-conditioned, recovering $\mathbf{x}$ from $\mathbf{y}$ becomes an ill-posed problem. A standard approach is therefore to introduce prior structural constraints through regularization,

\begin{equation}
    \hat{\mathbf{x}} = \arg\min_{\mathbf{x}} \frac{1}{2}\|\mathbf{y}-\mathbf{A}\mathbf{x}\|_2^2 + \lambda \mathcal{R}(\mathbf{x}).
\end{equation}

This formulation appears in many classical signal processing problems. For example, inverse blur problems are commonly solved using Tikhonov regularization, which leads to the Wiener deconvolution solution~\cite{Tikhonov1977, Engl1996, Vogel2002, Hansen2010}. More generally, many inverse reconstruction tasks can be understood as variations of the same regularized inverse problem, differing primarily in the forward operator $\mathbf{A}$ and the nature of the information bottleneck, as summarized in Table~\ref{tab:tasks}.

\subsection{Sparse Linear Regression}

Many natural signals are not sparse in their raw representation but become sparse under an appropriate transform basis. Let $\mathbf{x}=\mathbf{\Psi}\mathbf{w}$ denote a representation of the signal in a transform domain, where $\mathbf{w}$ is a coefficient vector.

Substituting this representation into the forward model yields
\begin{equation}
    \mathbf{y} = \mathbf{A}\mathbf{\Psi}\mathbf{w}.
\end{equation}
Defining the effective sensing matrix

\begin{equation}
    \mathbf{\Theta} = \mathbf{A}\mathbf{\Psi},
\end{equation}
the reconstruction problem becomes sparse linear regression,

\begin{equation}
    \hat{\mathbf{w}} = \arg\min_{\mathbf{w}} \frac{1}{2} \| \mathbf{y}-\mathbf{\Theta}\mathbf{w} \|_2^2 + \lambda \mathcal{R}(\mathbf{w}).
\end{equation}
The reconstructed signal is then given by

\begin{equation}
    \hat{\mathbf{x}}=\mathbf{\Psi}\hat{\mathbf{w}}.
\end{equation}

The central objective of sparse reconstruction is therefore accurate recovery of the support of the coefficient vector $\mathbf{w}$, i.e., identifying which coefficients correspond to meaningful signal components and which correspond to noise or irrelevant features.

\subsection{Regularizers as Prior Distributions}

Regularization terms $\mathcal{R}(\mathbf{w})$ can be interpreted as probabilistic prior assumptions on the coefficient distribution. Under a Bayesian interpretation, minimizing a regularized least-squares objective corresponds to maximum a posteriori (MAP) estimation with a particular prior distribution.

If no regularization is used, the implicit prior on $\mathbf{w}$ is flat.  
Different regularizers correspond to different priors:

\begin{itemize}

\item \textbf{Ridge regression / Tikhonov regularization}~\cite{ref:Ridge, Tikhonov1977}

\begin{equation}
    \min_{\mathbf{w}} \frac{1}{2}\|\mathbf{y}-\mathbf{\Theta}\mathbf{w}\|_2^2 + \lambda \|\mathbf{w}\|_2^2
\end{equation}
corresponds to a Gaussian prior on coefficients.

\item \textbf{LASSO ($\ell_1$ regularization)}~\cite{Tibshirani1996, Park2008}

\begin{equation}
    \min_{\mathbf{w}} \frac{1}{2}\|\mathbf{y}-\mathbf{\Theta}\mathbf{w}\|_2^2 + \lambda \|\mathbf{w}\|_1
\end{equation}
corresponds to a Laplace (double-exponential) prior.

\item \textbf{Total variation}~\cite{ref:TV}

\begin{equation}
    \min_{\mathbf{x}} \frac{1}{2}\|\mathbf{y}-\mathbf{A}\mathbf{x}\|_2^2 + \lambda \|\nabla \mathbf{x}\|_1
\end{equation}
can be interpreted as imposing a Laplace prior on image gradients, encouraging piecewise-smooth reconstructions. This can be interpreted as assuming sparsity in the gradient transform, where $\nabla$ acts as the sparsifying operator.

\item \textbf{$\ell_0$ sparsity}~\cite{ref:NNG, ref:SpikeAndSlab, George1997Approaches, Kappen2014}

\begin{equation}
    \min_{\mathbf{w}} \frac{1}{2}\|\mathbf{y}-\mathbf{\Theta}\mathbf{w}\|_2^2 + \lambda \|\mathbf{w}\|_0
\end{equation}
corresponds to a spike-and-slab prior, which explicitly separates active and inactive coefficients.

\end{itemize}

Among these approaches, LASSO has become one of the most widely used sparsity-inducing regularizer due to its convex formulation and computational tractability.~\cite{Donoho2006, Candes2006, Tibshirani1996, Park2008} However, LASSO performs continuous coefficient shrinkage and therefore introduces bias in coefficient magnitudes. More importantly, it does not explicitly separate relevant and irrelevant variables, which can limit support recovery accuracy in strongly sparse settings.

The ideal formulation for exact variable selection is therefore $\ell_0$ sparsity, which directly enforces a binary distinction between active and inactive coefficients. However, the resulting optimization problem is non-convex and generally NP-hard, motivating approximate approaches.

\subsection{Variational Garrote}

The Variational Garrote (VG)~\cite{Kappen2014} provides a tractable approximation to $\ell_0$ sparsity using latent binary selection variables. Let $s_i\in\{0,1\}$ denote binary gating variables controlling coefficient activity. The regression model becomes

\begin{equation}
    y_\mu = \sum_{i=1}^{N} w_i s_i X_{i\mu} + \xi_\mu ,
\end{equation}
where $\xi_\mu \sim \mathcal{N}(0,\beta^{-1})$.

Sparsity is imposed through a Bernoulli prior

\begin{equation}
    p(s_i|\gamma) = \frac{e^{\gamma s_i}}{1+e^\gamma},
\end{equation}
where $\gamma$ controls the expected sparsity level.

Exact inference over the binary variables is intractable.  
VG therefore employs a mean-field variational approximation

\begin{equation}
    q(\mathbf{s}) = \prod_i q_i(s_i),
\end{equation}
with activation probabilities $m_i = q(s_i=1)$. The resulting variational objective is the free energy

\begin{equation}
    F(\mathbf{w},\mathbf{m}) = \beta E_{\mathrm{rec}} + \Omega_{\mathrm{prior}} - H_{\mathrm{entropy}} - \ln \frac{\beta}{2\pi}.
\end{equation}

Following \cite{Soh2025}, the noise amplitude (inverse temperature) $\beta$ can be analytically optimized by tracing it out at the minimum of the reconstruction loss, leading to an effective objective proportional to $\log E_{\mathrm{rec}}$. The reconstruction energy is
\begin{align}
    E_{\mathrm{rec}} &= \frac{1}{2} \sum_{\mu} \left( y_\mu - \sum_i w_i m_i X_{i\mu} \right)^2 \nonumber \\
    &+ \frac{1}{2} \sum_{\mu}\sum_i m_i(1 -m_i) w_i^2 X_{i\mu}^2 ,
\end{align}
where the second term arises from the variance of the latent gating variables. The prior and entropy terms are

\begin{equation}
    \Omega_{\mathrm{prior}}=-\gamma\sum_i m_i ,
\end{equation}

\begin{equation}
    H_{\mathrm{entropy}} = -\sum_i \left[ m_i\log m_i + (1-m_i)\log(1-m_i) \right]. 
\end{equation}

Optimization of this free energy yields self-consistent updates for $\mathbf{w}$ and $\mathbf{m}$. Because coefficient magnitude and support selection are decoupled, VG reduces the shrinkage bias inherent to LASSO while approximating the spike-and-slab prior associated with $\ell_0$ sparsity.

\subsection{Connection to the Inverse Problems}

The inverse problems studied in this paper -- including signal resampling, denoising, and sparse-view CT reconstruction -- can all be expressed within this sparse regression framework. Each task differs in the forward operator $\mathbf{A}$ and the type of information bottleneck, but shares the same underlying reconstruction principle ({Table~\ref{tab:tasks}).

This perspective emphasizes that the effectiveness of a regularizer depends on how well its implicit prior matches the true coefficient distribution. In the experiments that follow we therefore compare LASSO and VG in several inverse reconstruction tasks to examine how different sparsity priors affect reconstruction accuracy and generalization performance.

\section{Experiments}

\begin{figure}
    \centering
    \includegraphics[width=\columnwidth]{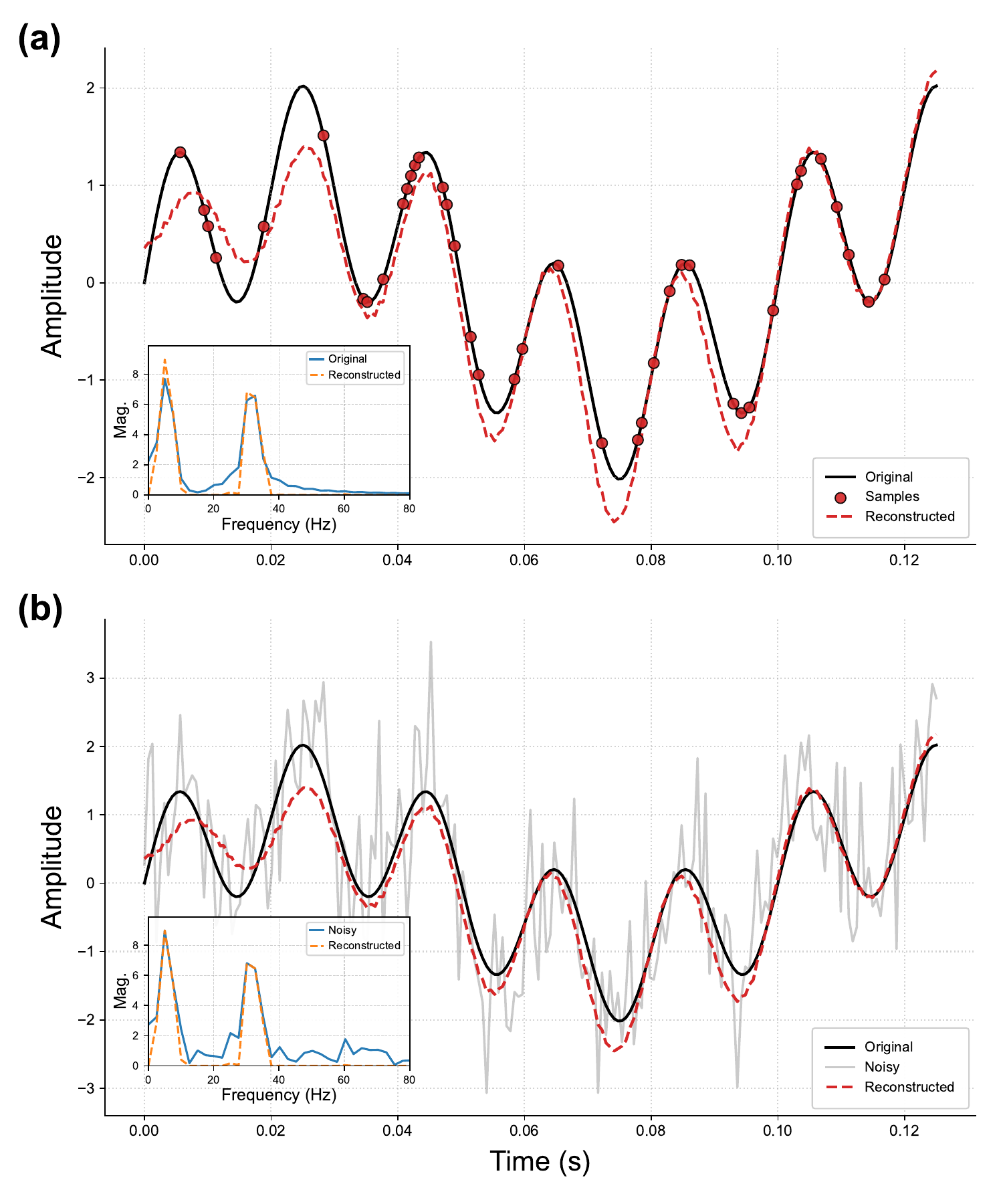}
    \caption{\textbf{Conceptual illustration of sparse reconstruction tasks.}
    (a) Signal resampling: a synthetic sinusoidal signal (black) is observed through random subsampling (red markers) and reconstructed using sparsity-based regularization (red dashed line). The inset shows the magnitude of the discrete cosine transform (DCT) spectrum. (b) Signal denoising: a clean signal (black) corrupted by additive Gaussian noise (gray) and its reconstructed estimate (red dashed line). The inset shows the corresponding DCT spectra of the noisy and reconstructed signals.}
    \label{fig:conceptual}
\end{figure}

This study adopts a unified experimental framework to compare sparsity-inducing priors across multiple inverse problems. In each experiment, a ground-truth signal or image is passed through a forward operator that imposes an \emph{information bottleneck} -- subsampling, noise corruption, or tomographic projection -- and reconstruction is performed by minimizing a regularized objective while sweeping the regularization strength over a wide range.

A wide variety of regularizers exist for inverse problems, including $\ell_2$ (Tikhonov)~\cite{Tikhonov1977} and total variation~\cite{ref:TV}. Here we focus on two representative and widely used sparsity-promoting approaches: the convex $\ell_1$ baseline (LASSO)~\cite{Tibshirani1996} and the Variational Garrote (VG)~\cite{Kappen2014}, which provides a probabilistic $\ell_0$-like estimator via latent gates. This choice isolates the practical differences between Laplace-type sparsity and spike-and-slab--like sparsity under controlled information bottlenecks.

\paragraph{Training protocol.}
All reconstructions are optimized to minimize the fitness loss with regularization on the regression coefficients. We consistently use the AdamW~\cite{AdamW} optimizer with initial learning rate 0.3, and a \texttt{ReduceLROnPlateau} scheduler until the learning rate reaches $10^{-5}$ (early stop), with a maximum of 50,000 iterations. For sound signal experiments, we use a batch of 100 independent realizations (masks or noise) to stabilize optimization. All parameters are initialized with small i.i.d. Gaussian noise.

\paragraph{Comparing regularization strengths across models.}
Because sparsity hyperparameters (e.g., $\lambda$ for LASSO vs.\ $\gamma$ for VG) are not directly commensurate across methods, comparing regularization strengths on an equal footing is difficult. We therefore rely on the bias–variance trade-off reflected in the train–generalization error curve. As regularization increases, the training error rises, whereas the generalization error initially decreases and then increases when regularization becomes excessive. Because stronger regularization increases the training error, the training error serves as an empirical proxy for the regularization strength. For each bottleneck setting, we sweep the hyperparameters, compute the corresponding generalization error, and define the optimal performance of each method as the minimum generalization error attained during the sweep. This provides a model-agnostic comparison of sparsity priors under each information bottleneck.


\subsection{Datasets}

We evaluate sparse reconstruction on two representative data modalities: sound signals and images. These domains were selected because both exhibit structural properties that make sparsity-based inverse reconstruction meaningful, while still representing fundamentally different signal characteristics.

Many other one-dimensional signals--such as financial time series or natural language sequences--do not possess a natural sparse representation in a fixed transform domain. Such signals typically require domain-specific generative models rather than generic sparse inverse methods. Because the goal of this work is to evaluate sparsity-driven reconstruction, we therefore restrict our experiments to domains where sparsity assumptions are physically or statistically justified.

Beyond these domains, higher-dimensional signals such as video or multidimensional sensor arrays could also be treated within the regularized inverse problem framework. For example, sparse reconstruction could be applied to recover temporal events or missing measurements in spatiotemporal signals. However, such settings introduce additional modeling complexities (e.g., motion dynamics and multi-axis sparsity structures). To maintain experimental clarity, we restrict our study to sound and image reconstruction tasks.

\paragraph{Sound signals.}

Acoustic signals generated by musical instruments typically exhibit strong harmonic structure. A sustained tone consists of a fundamental frequency and a limited number of harmonics, making the signal highly sparse in the Fourier or cosine transform domain. This property makes audio signals a natural testbed for transform-domain sparse reconstruction.
We consider two datasets representing controlled and real-world scenarios:

\begin{itemize}

\item \textbf{Synthetic sinusoidal signal.}
To provide a controlled benchmark with exact transform sparsity, we generate a synthetic signal consisting of two harmonic components,

\begin{equation*}
y(t)=\sin(1392\pi t)+\sin(3264\pi t), \quad t\in[0,1/8].
\end{equation*}

The signal is sampled with a unit sampling interval and therefore contains only two primary frequency components. This signal is strictly sparse in the cosine transform domain and serves as an idealized benchmark for evaluating sparse recovery behavior.

\item \textbf{Real-world audio signal.}
To evaluate reconstruction on real acoustic data, we use recordings from the TinySOL dataset \cite{TinySOL2020}, which contains high-quality isolated recordings of orchestral instruments. From this dataset, we select a sustained flute tone with pitch G5 (approximately 784\,Hz). To avoid transient effects from the attack phase of the instrument, we extract a segment from the steady-state portion of the recording starting at approximately 2.5\,s.

\end{itemize}

The audio signal is resampled to a sampling rate of 16\,kHz. For reconstruction experiments, signals are discretized to length $N=2500$, centered to zero mean, and normalized to unit variance. Sparse reconstruction is performed in the cosine transform domain using the discrete cosine transform (DCT-II) implemented in \texttt{scipy} with normalization parameter \texttt{norm=`ortho'}.

\paragraph{Images.}

Unlike sound signals, natural images are generally not strictly sparse in the frequency domain. However, they exhibit strong spatial structure: pixels belonging to the same object or region tend to share similar intensities, producing piecewise-smooth regions separated by relatively sparse boundaries. This structural property enables sparse or structured regularization to effectively constrain image reconstruction problems.
To evaluate imaging reconstruction tasks, we consider images from four datasets widely used in tomographic reconstruction studies:

\begin{itemize}

\item \textbf{Shepp--Logan phantom} \cite{Shepp1974}. This synthetic phantom is a classical benchmark in CT reconstruction research. The image consists of a collection of ellipses representing simplified anatomical structures, providing a controlled dataset with clearly defined boundaries.

\item \textbf{LIDC-IDRI (Lung CT)} \cite{Armato2011}. This dataset contains clinical lung CT scans with complex anatomical structures and heterogeneous textures, providing realistic medical imaging data for reconstruction evaluation. 

\item \textbf{BraTS (Brain MRI)} \cite{Menze2015}. The BraTS dataset contains brain MRI scans commonly used in tumor segmentation research. These images include smooth anatomical regions as well as localized structural irregularities.

\item \textbf{Walnut dataset} \cite{Hamalainen2015}. This dataset consists of high-resolution CT scans of walnuts and is widely used as a benchmark for sparse-view CT reconstruction due to its intricate internal structures.

\end{itemize}

All images are resized to $512\times512$, normalized to the range $[0,1]$, and restricted to a circular field-of-view to match the acquisition geometry used in tomographic reconstruction. For imaging experiments we simulate parallel-beam CT acquisition using the Radon transform. Projection data (sinograms) are generated with a detector count of 512, matching the image width, with parallel-beam geometry is used throughout all experiments. To avoid corner artifacts caused by square image boundaries, reconstructions are restricted to a circular field-of-view mask with diameter equal to image dimension.

These datasets therefore allow sparse reconstruction to be evaluated across both strictly sparse transform-domain signals (audio) and spatially structured signals (images), providing complementary test cases for sparse regularization.

\begin{figure*}
    \centering
    \includegraphics[width=\textwidth]{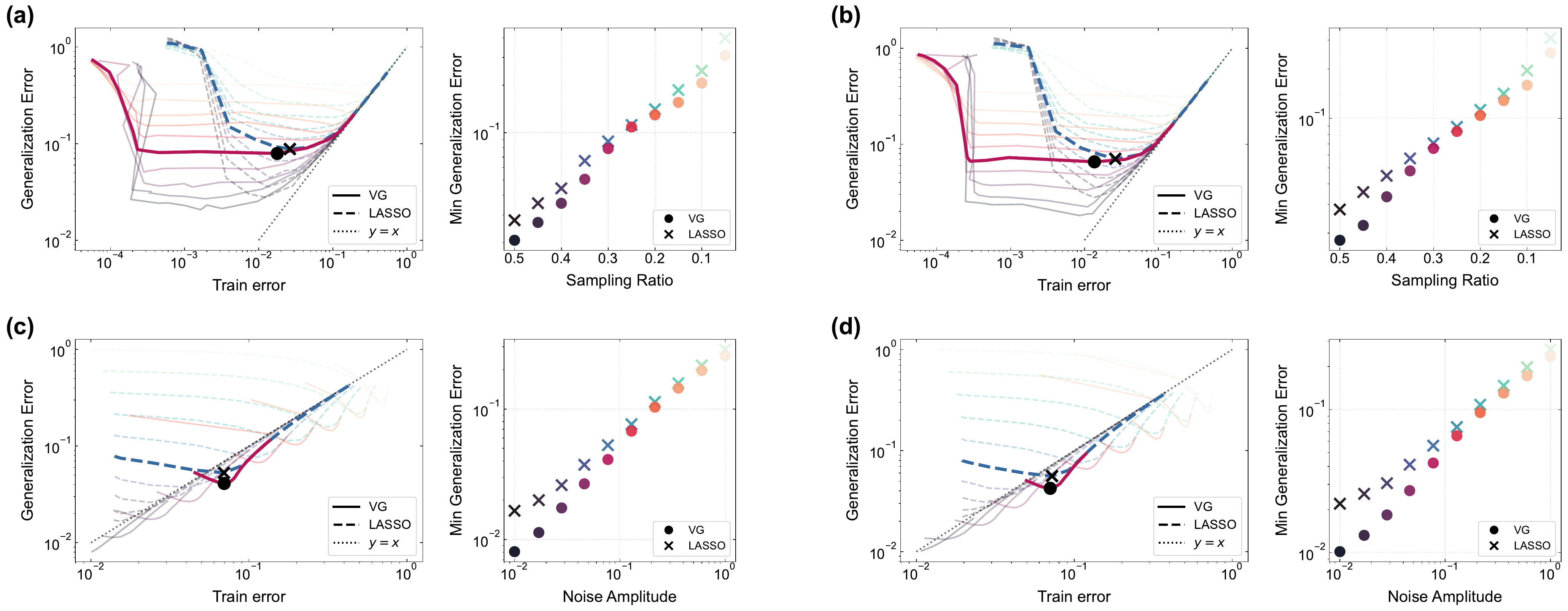}
    \caption{\textbf{Train–generalization ($\mathcal{E}_{\mathrm{train}}$ vs. $\mathcal{E}_{\mathrm{gen}}$) error curves for LASSO and Variational Garrote (VG).} Top row: signal resampling experiments for the synthetic signal (a) and real flute signal (b). Bottom row: signal denoising experiments for the synthetic (c) and real (d) datasets. For each task, the left plot shows the relationship between training error and generalization error on log–log scales. Solid lines denote VG and dashed lines denote LASSO. Curve colors indicate different sampling ratios (top) or noise amplitudes (bottom). Markers indicate the hyperparameter setting achieving the minimum generalization error. The dotted diagonal line represents $\mathcal{E}_{\mathrm{train}}=\mathcal{E}_{\mathrm{gen}}$. The adjacent right plot explicitly shows this minimum generalization error as a function of the sampling ratio or noise amplitude.}
    \label{fig:1d_performance_comparison}
\end{figure*}

\subsection{Signal Resampling}
We first evaluate reconstruction under random subsampling of sound signals. As illustrated in Fig.~\ref{fig:conceptual}(a), measurements are obtained by applying a random binary mask to the time-domain signal. All experiments are performed on the same underlying signal while varying the observation mask.

Let $\Omega$ denote the set of observed indices and $\Omega^c$ the set of missing indices. The sampling ratio is given by
\begin{equation}
    R=\frac{|\Omega|}{N}.
\end{equation}
To quantify the reconstruction performance, we report the relative errors on both the observed and missing samples. Because the signals are centered to zero mean, these relative $\ell_2$ errors correspond exactly to the root mean square error (RMSE) normalized by the standard deviation of the target samples:
\begin{equation}
    \mathcal{E}_{\rm train} = \frac{\|\mathbf{x}_{\Omega}-\hat{\mathbf{x}}_{\Omega}\|_2}{\|\mathbf{x}_{\Omega}\|_2}, \quad
    \mathcal{E}_{\rm gen}   = \frac{\|\mathbf{x}_{\Omega^c}-\hat{\mathbf{x}}_{\Omega^c}\|_2}{\|\mathbf{x}_{\Omega^c}\|_2},
\end{equation}
where $\hat{\mathbf{x}}$ is the signal reconstructed from the masked observations. Reconstruction is performed in the DCT domain, $\mathbf{x}=\Psi\mathbf{s}$, via sparse regression over $\mathbf{s}$.

We vary $R$ from $5\%$ to $50\%$ in increments of $5\%$. For each $R$, masks are sampled uniformly without replacement. Hyperparameters are selected via grid search: LASSO $\lambda \in [5 \times 10^{-4}, 5]$ (log scale) and VG $\gamma \in [-15,-1]$. To ensure statistical reliability, each configuration is repeated over $100$ independent random masks, and the average metrics are reported.

Figure~\ref{fig:1d_performance_comparison}\,(a,\,b) shows the train--generalization error curves for synthetic sinusoidal signals and real-world audio signals. Both methods exhibit the characteristic bias--variance tradeoff: points near $\mathcal{E}_{\rm gen}\approx\mathcal{E}_{\rm train}$ correspond to strongly regularized (underfitting) solutions, where the reconstruction fails even on the observed samples. As regularization weakens, both errors decrease until reaching a minimum generalization error, beyond which overfitting emerges. In the overfitting regime, the reconstruction matches observed samples extremely well (low train error) while severely deviating on missing samples (increased generalization error).

Notably, VG exhibits abrupt transitions in train error as $\gamma$ is swept. This behavior is consistent with gate-driven support selection: as dominant spectral peaks enter the effective ``slab'' region of the smoothed spike-and-slab prior, entire frequency components become abruptly activated or deactivated, producing phase-transition--like changes in the fit. In contrast, LASSO shrinks coefficients continuously, yielding smoother error trajectories.

Insets report the minimum generalization error (MGE) versus the sampling ratio. VG consistently achieves a lower MGE than LASSO across both datasets, with the most significant gains observed in low-sampling regimes ($R<0.2$), where performance is heavily dominated by correct support recovery under severe information bottlenecks.

\subsection{Signal Denoising}
We next evaluate the robustness to additive white Gaussian noise (AWGN). As illustrated in Fig.~\ref{fig:conceptual}(b), noisy observations are generated by
\begin{equation}
    \mathbf{x}_{\rm noisy}=\mathbf{x}_{\rm clean} + \alpha\boldsymbol{\epsilon}, \quad \boldsymbol{\epsilon}\sim\mathcal{N}(0,I).
\end{equation}
Reconstruction is again performed in the DCT domain, $\mathbf{x}=\Psi\mathbf{s}$, by sparse regression over $\mathbf{s}$.

Following the same normalized RMSE rationale, the train error measures the relative $\ell_2$ fit to the noisy observation,
\begin{equation}
    \mathcal{E}_{\rm train} = \frac{\|\mathbf{x}_{\rm noisy}-\hat{\mathbf{x}}\|_2}{\|\mathbf{x}_{\rm noisy}\|_2},
\end{equation}
while generalization error measures recovery of the clean signal,
\begin{equation}
    \mathcal{E}_{\rm gen} = \frac{\|\mathbf{x}_{\rm clean}-\hat{\mathbf{x}}\|_2}{\|\mathbf{x}_{\rm clean}\|_2}.
\end{equation}

We sweep the noise amplitude $\alpha$ over 10 logarithmically spaced values between $10^{-2}$ and $10^{0}$ (corresponding to approximately 40\,dB to 0\,dB SNR after signal normalization). For robust evaluation, results are averaged over $50$ independent noise realizations. The hyperparameter search ranges follow those used in the resampling experiments.

Figure~\ref{fig:1d_performance_comparison}\,(c,\,d) displays the train--generalization curves. As in the resampling task, both methods exhibit a bias--variance tradeoff: overly strong regularization underfits the data (errors are both high and approach the diagonal), whereas weak regularization overfits the noisy observation (train error becomes small while generalization error increases). Unlike resampling, however, the overfitting regime does not drive $\mathcal{E}_{\rm gen}$ toward unity; instead, the generalization error saturates at a noise-dominated floor determined primarily by $\alpha$, reflecting the intrinsic information loss in the observation.

Furthermore, VG does not exhibit the same abrupt train-error jumps observed in resampling. This is consistent with the fact that noise blurs the effective spectral support, meaning that small hyperparameter changes no longer trigger the discrete activation of clean components. Insets report MGE versus $\alpha$: VG achieves a lower minimum generalization error across the entire noise range, with the clearest advantages emerging at low-to-moderate noise levels where the sparse harmonic structure remains identifiable despite the corruption.

\subsection{Sparse-view CT Reconstruction}

To evaluate imaging reconstruction we consider sparse-view computed tomography (CT), where measurements are line integrals modeled by the Radon transform~\cite{Radon1917}. In discretized form, an image $\mathbf{x}\in\mathbb{R}^{N^2}$ is mapped to a sinogram $\mathbf{y}$ through a linear projection operator $\mathbf{A}$, yielding a regularized linear inverse problem in the same class studied above. Forward and adjoint operators are implemented using \texttt{torch\_radon}~\cite{torch_radon}.

Sparse-view CT reduces the number of projection angles $K$, making the inverse problem severely underdetermined. For each image we generate sinograms using $K$ uniformly spaced angles in $[0,\pi)$ with parallel-beam geometry and detector count 512, and vary $K\in\{10,20,\dots,120\}$. We compare LASSO and VG, and include filtered back-projection (FBP) as an analytical baseline. Hyperparameters are swept by grid search: $\lambda \in [10^{1},10^{3}]$ for LASSO and $\gamma \in [-10,0]$ for VG.

We reconstruct directly in pixel space to isolate the effect of coefficient priors without introducing additional sparsifying transforms. While generic natural images are not pixel-sparse, CT slices under a circular field-of-view often contain large low-attenuation background regions and relatively compact support, making pixel-domain sparsity a meaningful stabilizing baseline prior. The optimization objective is defined in the sinogram domain via data fidelity, while performance is evaluated in the image domain using MSE, reflecting the end goal of accurate attenuation-map recovery.

Figure~\ref{fig:qualitative_reconstruction} shows qualitative reconstructions at $K=40$. FBP exhibits severe streak artifacts under angular undersampling, whereas both sparse regularizers suppress these artifacts. Figure~\ref{fig:quantitative_robustness} reports MSE versus $K$ (mean$\pm$std over 10 trials). Across most angular resolutions, reconstruction error follows the trend FBP $>$ LASSO $>$ VG, and VG typically shows smaller variability, indicating improved stability under initialization and measurement variability. Empirically, VG tends to reduce bulk error in homogeneous regions while occasionally producing slightly weaker boundary sharpness, suggesting potential benefits from combining VG with complementary edge-preserving priors such as gradient-domain sparsity or total variation.

\begin{figure*}
    \centering
    \includegraphics[width=\textwidth]{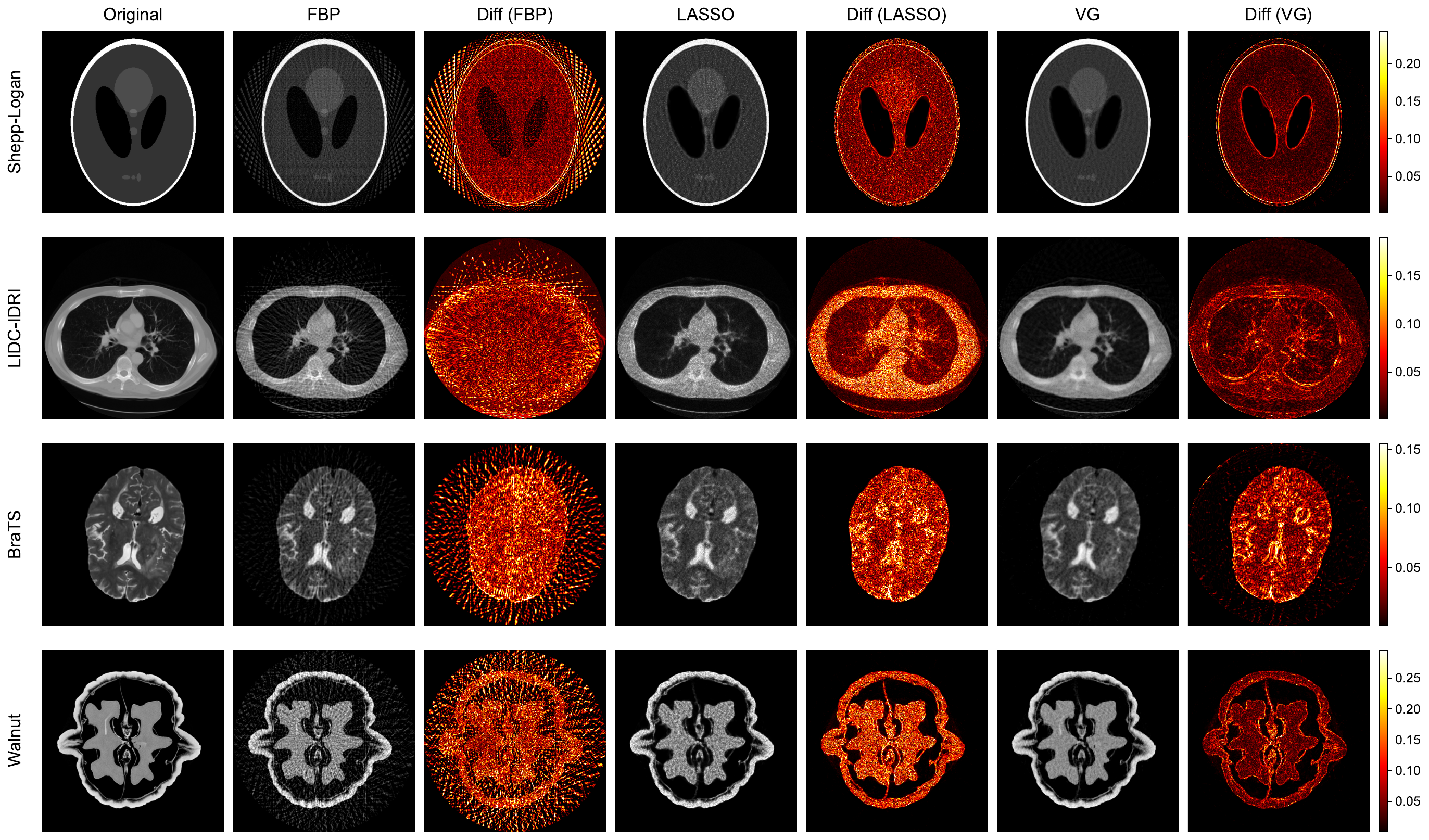}
    \caption{\textbf{Qualitative comparison of sparse-view CT reconstructions ($K=40$ projection angles).} Rows correspond to four datasets: Shepp–Logan phantom, LIDC-IDRI chest CT, BraTS brain MRI, and Walnut CT.  Columns show the ground truth image, reconstructions obtained with filtered back-projection (FBP), LASSO, and the Variational Garrote (VG), together with their corresponding absolute error maps (``Diff'').  Brighter colors in the difference maps indicate larger reconstruction errors.
    }
    \label{fig:qualitative_reconstruction}
\end{figure*}

\begin{figure*}
    \centering
    \includegraphics[width=\textwidth]{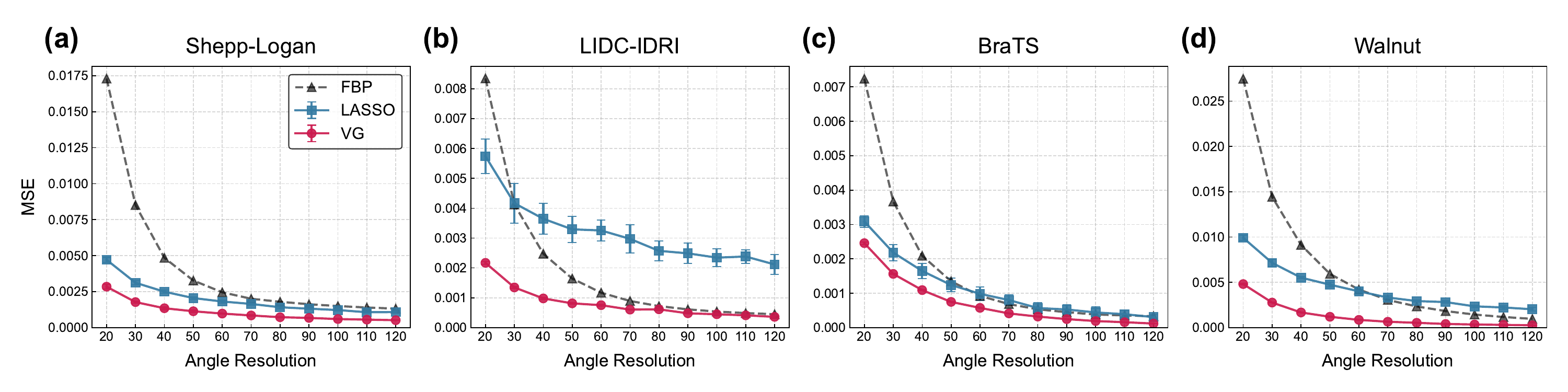}
    \caption{\textbf{Reconstruction error versus angular resolution in sparse-view CT.} Mean squared error (MSE) between reconstructed and ground-truth images is shown as a function of the number of projection angles $K$ for four datasets: (a) Shepp–Logan phantom, (b) LIDC-IDRI chest CT, (c) BraTS brain MRI, and (d) Walnut CT. Filtered back-projection (FBP), LASSO, and Variational Garrote (VG) are compared. Each point corresponds to the reconstruction obtained with the hyperparameter that minimizes MSE for the given $K$. Markers show the mean over 10 trials and error bars denote standard deviation.
    }
    \label{fig:quantitative_robustness}
\end{figure*}

\subsection{Experimental Summary}

Across both signal and image reconstruction tasks several consistent empirical patterns emerge. First, both LASSO and VG exhibit the characteristic bias–variance tradeoff as the regularization strength varies. For each reconstruction task the optimal performance is achieved at an intermediate regularization level corresponding to the minimum generalization error.

Second, VG tends to achieve lower minimum generalization error in strongly underdetermined regimes. This effect is particularly visible when the information bottleneck is severe, such as low sampling ratios in signal resampling or limited angular resolution in sparse-view CT. In such settings, accurate recovery of the sparse support becomes the dominant factor determining reconstruction quality.

Third, the imaging experiments reveal qualitative differences between the two sparsity priors. VG typically reduces reconstruction error across large homogeneous regions while occasionally producing slightly weaker boundary sharpness. This behavior suggests that combining VG with complementary edge-preserving priors (e.g., gradient-domain sparsity or total variation regularization) may further improve reconstruction quality in imaging tasks.

\section{Conclusion}

In this work we investigated sparse inverse reconstruction from the perspective of sparse linear regression. By expressing signal resampling, denoising, and sparse-view CT reconstruction within a unified regularized inverse problem framework, we examined how different sparsity-inducing priors influence reconstruction behavior under controlled information bottlenecks.

Our experiments compared the widely used $\ell_1$ regularization (LASSO) with the Variational Garrote (VG), a probabilistic $\ell_0$-like sparse estimator that introduces latent gating variables to separate coefficient magnitude from support selection. Across the tasks considered, both approaches exhibited the characteristic bias–variance tradeoff as the regularization strength varied, with optimal reconstruction performance occurring at an intermediate level of regularization.

A consistent observation across experiments is that VG often achieves lower minimum generalization error when the inverse problem is strongly underdetermined. This behavior can be interpreted through the probabilistic priors associated with each regularizer. In sparse regression, regularization corresponds to assumptions about the coefficient distribution: LASSO induces a Laplace (double-exponential) prior, while $\ell_0$-type formulations correspond more closely to spike-and-slab priors that explicitly separate active and inactive coefficients. When the underlying coefficient distribution is strongly sparse, spike-and-slab–like models provide a closer approximation to the data-generating process, which can lead to improved support recovery and lower reconstruction error. Importantly, this effect appears across both signal and imaging tasks, suggesting that the advantage arises primarily from prior–data alignment rather than domain-specific properties.

The imaging experiments further suggest that different sparsity priors may influence reconstruction structure in distinct ways. In sparse-view CT, VG often improves fidelity in large homogeneous regions while occasionally producing slightly weaker boundary sharpness compared with LASSO. Since many downstream medical imaging tasks--such as segmentation or region-based analysis--depend on accurate recovery of homogeneous regions, improved bulk reconstruction fidelity may translate into practical advantages in such applications.

Although VG introduces additional latent variables and therefore increases computational complexity relative to LASSO, the overhead remains moderate in practice, primarily involving an additional set of gating variables. Unlike convex $\ell_1$ optimization, however, global convergence guarantees are generally not available, and optimization behavior may depend on initialization and training schedules.

Beyond the experiments presented here, the results suggest several directions for future work. First, the same prior-selection principle may extend to nonlinear models such as deep neural networks~\cite{Louizos2017}. For example, sparsity-inducing priors similar to VG could be applied to the final or penultimate layer weights of deep architectures, allowing structured sparsity assumptions to guide representation learning while preserving nonlinear modeling capacity.

Second, the prior–data alignment perspective naturally extends to other structured regularizers. In imaging problems, total variation~\cite{ref:TV} regularization implicitly assumes sparsity in the gradient domain. Under this view, reconstruction quality depends on how closely the empirical gradient distribution matches the assumed prior. Extending spike-and-slab–like priors to gradient-domain representations may therefore provide improved reconstruction in piecewise-smooth images.

Finally, the framework can be generalized to broader classes of inverse problems. Many signal processing tasks--such as deconvolution, Wiener filtering variants, and other regularized inverse problems--share the same linear operator structure studied here. In more challenging settings where the forward operator itself must be estimated, such as blind inverse problems~\cite{Levin2009} or learnable forward models, sparsity priors may play an even more important role in stabilizing reconstruction.

Overall, these results highlight the importance of viewing regularization not only as an optimization tool but also as an explicit probabilistic assumption about parameter distributions. Understanding how these priors align with the structure of real data provides a principled path toward improving reconstruction performance across a wide range of inverse problems.

\subsection*{Acknowledgement}
This work was supported by the National Research Foundation of Korea (NRF) grant (Grant No. 2022R1A2C1006871) (J.J.).

\bibliographystyle{model1-num-names}

\bibliography{reference}

\end{document}